%% file: main.tex
\documentclass[10pt,twocolumn,letterpaper]{article}

\usepackage{wacv}
\usepackage{times}
\usepackage{epsfig}
\usepackage{graphicx}
\usepackage{amsmath}
\usepackage{amssymb}
\usepackage{booktabs}
\usepackage{multirow}

\usepackage{color}
\usepackage{comment}
\usepackage{amsmath,amssymb} 
\newtheorem{definition}{Definition}

\usepackage{subcaption}
\usepackage{float}

\newcommand\Red[1]{\textcolor{red}{#1}}
\newcommand\Blue[1]{\textcolor{blue}{#1}}

\wacvfinalcopy 

\ifwacvfinal
\usepackage[breaklinks=true,bookmarks=false]{hyperref}
\else
\usepackage[pagebackref=true,breaklinks=true,colorlinks,bookmarks=false]{hyperref}
\fi

\begin{document}

\title{EVAL: Explainable Video Anomaly Localization}

\author{Ashish Singh\\
University of Massachusetts Amherst\\
{\tt\small asingh@cs.umass.edu}
\and
Michael J. Jones\\
Mitsubishi Electric Research Laboratories\\
{\tt\small mjones@merl.com}
\and
Erik Learned-Miller\\
University of Massachusetts Amherst\\
{\tt\small elm@cs.umass.edu}
}

\maketitle
\thispagestyle{empty}

\input{abstract.tex}
\input{body.tex}


{\small
\bibliographystyle{ieee_fullname}
\bibliography{egbib}
}

\input{body_supplemental.tex}

\end{document}

%% file: abstract.tex
\begin{abstract}

We develop a novel framework for single-scene video anomaly localization that allows for human-understandable reasons for the decisions the system makes.  We  first learn general representations of objects and their motions (using deep networks) and then use these representations to build a high-level, location-dependent model of any particular scene. This model can be used to detect anomalies in new videos of the same scene.  Importantly, our approach is {\em explainable} -- our high-level appearance and motion features can provide human-understandable reasons for why any part of a video is classified as normal or anomalous.   We conduct experiments on standard video anomaly detection datasets (Street Scene, CUHK Avenue, ShanghaiTech and UCSD Ped1, Ped2) and show significant improvements over the previous state-of-the-art.

\end{abstract}

%% file: body.tex
\section{Introduction}
\label{sec:intro}
We are interested in the problem of spatio-temporal localization of anomalous activities in videos of a given scene. Informally, anomalous activities are  events that differ from those typically observed in a scene, such as a cyclist riding through an indoor shopping mall \cite{ramachandra2020survey}.  Like many other papers on anomaly detection, this work addresses the setting in which we have access to an initial set of videos that are used to define the typical, or `nominal' activities in a particular scene. Such a situation naturally arises in  surveillance and monitoring tasks \cite{saligrama2010video}, where it is easy to collect nominal data, but it is not practical to collect a representative set of possible anomalies for a scene.
Thus, the problem set-up is as follows: provided with a set of videos of a scene which do not contain any anomalies, (called the {\em nominal set}), the goal is to detect any events in a test video from the same scene that differ substantially from all events in the nominal set \cite{mahadevan2010anomaly, ramachandra2020survey}.  In defining anomaly detection, it is important to consider the role of {\em location}.
In real-world surveillance scenarios, an event may be normal in one location but anomalous in another. For example, a car driving on a road is typically not anomalous, while one driving on a sidewalk typically is. In view of this, we adopt the following definition \cite{ramachandra2020survey}.

\begin{definition}
\vspace{-8pt}
 An {\it anomaly} is any spatio-temporal region of test video that is significantly different from all of the nominal video in the same spatial region. 
 \vspace{-10pt}
\end{definition}

Unlike most recent work in anomaly detection, a key goal of our work is to produce not only a set of anomalies, but a simple and clear explanation for what makes them anomalous. 
We are motivated by how people tasked with watching video from a stationary surveillance camera would detect an unusual incident.
While monitoring a scene, we expect a person to note the types of objects seen (people, buildings, cars) and the motions of those objects (walking east or west on a sidewalk, driving northwest on the street) to characterize the given scene.  The person could then notice an anomaly when the objects or motions do not match what has been seen before.  The person could also explain why something was anomalous.

We design our video anomaly detection system using this sketch of how a human would solve the problem as motivation.  We want to use deep networks to give a high-level understanding of the objects and motions occurring in a scene. By 'high-level', we mean at the level of whole objects and not at the level of pixels or edges.
To do this, we train deep networks that take a spatio-temporal region of video (which we call a {\em video volume}) as input and output attribute vectors representing the object classes, the directions and speeds of motion and the fraction of stationary pixels (which gives information on the sizes of moving objects) occurring in a spatio-temporal region.   The feature vectors from the penultimate layers of these deep networks yield high-level representations, or {\em embeddings}, of the appearance and motion content of each video volume. Ten frames are used for video volumes in our experiments.

Unlike many other recent works, we do {\em not} learn new embedding functions (i.e. networks) for new scenes. We use the same embedding networks for every environment. Instead, to characterize the nominal video for a new scene, we store a representative set of all the embeddings found in the nominal video. That is, for every video volume in the nominal video of a new scene, we calculate our representations of appearance, motion direction, speed, and background fraction. We then reduce this set of embeddings to a smaller set which we call {\it exemplars} by removing redundant embeddings. This results in a compact, accurate, and location-dependent model of the nominal data in a new scene.
Since there is no training of deep networks for each new environment, modeling a new environment is `lightweight' compared to many other methods, making it efficient to model new scenes.  Our exemplar model also allows very efficient updating if new nominal video is introduced.  This is a crucial property for video anomaly detection methods because, in practice, it is unrealistic to assume that the initial nominal video covers every possible normal change.  New nominal video will occasionally need to be added, making it critical that models are easy to extend.

Given test video of the same scene, we compute our high-level features for each video volume.  We then compare these to the exemplars stored in the nominal model at the corresponding spatial region.  Any test feature with a high distance to \textit{every} nominal exemplar for that region is considered anomalous.
Because the feature vectors map to human-interpretable attributes, these attributes can be used to give human-understandable explanations for why our system labeled video volumes as normal or anomalous.
We define a method as 'explainable' if it can give human-understandable reasons for its decisions.  Details of how our system provides explanations are given in Section \ref{sec:explainability}.

In summary, we make the following key contributions:  1) We show that modeling scenes using high-level attributes leads to robust anomaly detection.
2) We introduce the idea of directly estimating high-level motion attributes from raw video volumes using deep networks.
3) We show how these high-level attributes also allow human-interpretable explanations.
4) Finally, we demonstrate an alternative to much of the previous work that is based on learning to reconstruct the nominal data.  Our alternative approach is practical since it does not require training deep networks for each new scene and allows for simple and efficient updates to a scene model given new nominal training data.


\section{Related Work}
\label{sec:related}

\begin{figure*}[]
    \centering
    \includegraphics[width=0.78\linewidth]{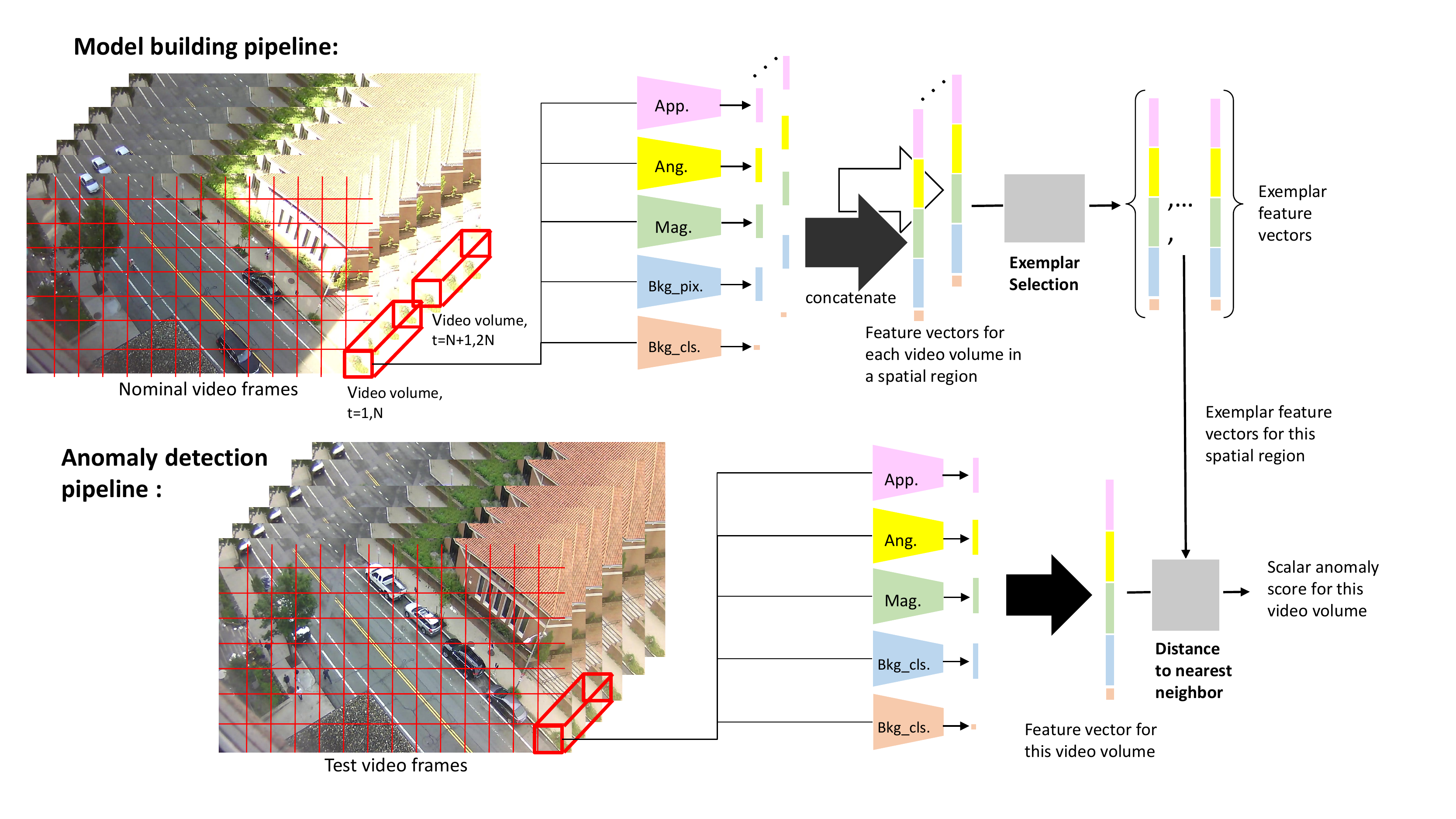}
    \vspace{-10pt}
    \caption{Our pipeline for building a location-dependent model of  nominal video and detecting anomalies in test video. During the model building phase, we extract a high-level representation of each video volume using our appearance and motion networks. Using the exemplar selection method, we select a representative subset of video volumes for a given spatial region. By comparing video volumes in test video to the exemplar set we can detect anomalies.
    }
    \label{fig:modelbuilding}
    \vspace{-10pt}
\end{figure*}

Most Video Anomaly Detection (VAD) methods can be analyzed in terms of their representation learning or their detection method.
\\
\textbf{Representation of Nominal Data:}  Early approaches to VAD \cite{adam2008robust, antic2011video, cong2013abnormal, li2013anomaly, mehran2009abnormal, saligrama2012video, wu2010chaotic} primarily relied on the usage of handcrafted features. This included features like spatio-temporal gradients \cite{lu2013abnormal, ionescu2019detecting}, histogram of gradients \cite{saligrama2012video, ma2015anomaly}, flow fields \cite{adam2008robust, antic2011video, wu2010chaotic, mehran2009abnormal},  histogram of flows \cite{saligrama2010video,saligrama2012video,cong2013abnormal}, dense trajectories \cite{ma2015anomaly, stauffer2000learning} and foreground masks \cite{antic2011video}. Recently, most authors have used deep learning for this task \cite{doshi2020any, doshi2020continual, hasan2016learning, hinami2017joint, ionescu2019object, liu2018future, luo2017revisit, ramachandra2020learning, ravanbakhsh2017abnormal, ravanbakhsh2018plug, smeureanu2017deep, wang2018abnormal, sabokrou2017deep, 10.1016/j.cviu.2016.10.010}. These methods either use a pretrained model \cite{smeureanu2017deep, ionescu2019detecting, hinami2017joint, ravanbakhsh2018plug, luo2017revisit, sabokrou2017deep} for feature extraction or train a model to specifically optimize for a particular task related to anomaly detection. These tasks can generally be categorized as either a variant of training an auto-encoder architecture to minimize the reconstruction error of nominal frames \cite{hasan2016learning,ionescu2019object,nguyen2019anomaly,chang2020clustering,lu2020few,liu2021hybrid}, a generative adversarial network (GAN) to model nominal frames \cite{liu2018future, lu2020few}, or future frame prediction given a sequence of nominal frames \cite{liu2018future,wang2021prediction}.  To further improve their performance, recent works have tried specialized architectures and training methodologies. Particularly \cite{dong2020dual, liu2021hybrid, park2020learning} transform their respective generative model by memory-based modules to memorize the normal prototypes in the training data. Most recently \cite{ristea2021self} proposed a module based on masked convolution and channel attention  to  reconstruct  a  masked  part  of  the  convolutional receptive field.  A key drawback of reconstruction and future frame prediction methods is that they do not generate interpretable features. It is not clear what aspects of the video make it difficult to reconstruct since there is no mapping to higher-level features as in our work.
Our work mostly aligns with methods utilizing pretrained models. However, unlike most of these approaches, we incorporate the output of our pretrained models to interpret model decisions. We additionally make sure that the predictions of our model generalize to a wide variety of scenes through our data generation and training procedures.  One high-level motion attribute that our method learns is similar to the histogram of flow feature used in some early work \cite{saligrama2010video,saligrama2012video,cong2013abnormal}.  However, instead of computing the histogram of flow from an optical flow field, our method learns a deep network to predict these features directly from RGB video volumes and then uses the network's learned feature embedding as the representation.
\\
\textbf{Detection Methods:} Most methods use either standard outlier detection methods \cite{chandola2009anomaly} as an external module or utilize the reconstruction strategy to predict anomalies. General methods of detection that most authors have used in the past include one-class SVM \cite{smeureanu2017deep, ionescu2019detecting, ma2015anomaly, xu2015learning}, nearest neighbour approaches \cite{ramachandra2020street, ramachandra2020learning, hinami2017joint, doshi2020any, doshi2021efficient}, and Probabilistic Graphical Models \cite{antic2011video}. In \cite{astrid2021synthetic, georgescu2021background}, pseudo-anomalous samples are used during training to improve discriminative learning.  In \cite{georgescu2021anomaly}, an object detector is used to focus on regions around objects and then networks are trained for various 'proxy' tasks (such as predicting the arrow of time) on the nominal data.  Thus, unlike our work, they require training deep nework models for each different scene.
Our work has some similarity to the work of \cite{ramachandra2020street,ramachandra2020learning} in that we also use an overlapping grid of spatial regions, build exemplar-based models and use nearest neighbors distances as anomaly scores.  The high-level features that we use are the biggest difference as compared to the pixel-based features used in their work.  Our high-level features allow for explainable models as well as much smaller models than theirs.
\\
\textbf{Explainable VAD:} 
Our work is similar in spirit to the work of \cite{hinami2017joint, wu2021explainable} with respect to providing explanations for detecting anomalies. In \cite{hinami2017joint}, the authors pre-train their feature extractor on public image datasets (MS-COCO and Visual Genome) to detect objects and predict their attributes and actions. They further use these predictions for 'event recounting' on VAD benchmarks. In \cite{wu2021explainable}, the authors utilize models pretrained for semantic segmentation, object classification, and multi-object tracking
and use the output of these models directly as their feature representation.

Despite these coarse similarities, virtually all of the details of our methods are different. Specifically, in \cite{hinami2017joint}, they rely on object proposals to find candidate anomalous regions which can lead to missed detections for objects not represented in their training data. Our action/motion classes are also very different - ours being more generic (direction distributions and speed of motion) while in \cite{hinami2017joint} are much more specific (bending, riding) and hence not applicable to a wide variety of scenarios.
The method of \cite{wu2021explainable} is specific to detecting and tracking pedestrians and is not a general video anomaly detection method.  Furthermore, unlike ours, their method does not spatially localize anomalies.

\section{Our Approach}

Our method consists of three distinct stages: high-level attribute learning, model building, and anomaly localization.  The high-level attribute learning stage is done only once and uses training samples that are independent of any video anomaly detection dataset.  The resulting deep networks learn general representations of object appearances and object motions which can then be used in the subsequent two stages to build a model of a specific scene and to localize anomalies in that scene, for a wide variety of surveillance scenarios. The outside data used to train our high-level attribute models are equivalent to the outside data used in various prior works on VAD; for example, the MS-COCO and Visual Genome data used to train the models of Hinami et al. \cite{hinami2017joint}, the outside data used to train object detectors in \cite{georgescu2021anomaly,doshi2021efficient,ionescu2019detecting,smeureanu2017deep}, as well as the many deep models pretrained on ImageNet and applied to VAD.  Our outside data is not used to build models of a scene.

\vspace{-2pt}
\subsection{\bf{High-Level Attribute Learning}}

For this stage, our main objective is to learn features that are \textbf{(a)} transferable across scenes and \textbf{(b)} interpretable. Given this motivation, we learn an object recognizer for our appearance model as well as regression networks for estimating the following motion attributes for a given video volume: the fraction of stationary pixels, the distribution of motion directions and the average speed of movement in each direction.  We also learn a classifier to indicate whether a video volume is stationary or not.  We will describe each of these deep networks in the following subsections.

To clarify our terminology, we use the term {\em 'high-level attribute'} to denote the object classes, histogram of motion directions, vector of motion speeds or fraction of stationary pixels which are the final outputs of the various deep networks that are learned.  The term {\em 'high-level feature'} denotes the feature vector from the penultimate layer of one of the deep networks.  A high-level feature can be mapped to a high-level attribute using the final layer of that network.
\vspace{-10pt}
\subsubsection{Appearance model}
We formulate the task of object recognition as a multi-label image classification problem as any given input image patch may contain more than one object class (or none). 

\textbf{Training Data:}
We are particularly interested in learning to recognize objects that have high likelihood of being present in outdoor scenes. To this end, we select the following 8 categories as our primary set of object classes : [Person, Car, Cyclist, Dog, Tree, House, Skyscraper, and Bridge]. For our formulation, we want the learned features to generalize across different domains. To achieve this, we construct our training dataset of images from multiple sources. 
We use labeled examples of each class (as well as background images containing none of the classes) taken from the CIFAR-10 \cite{krizhevsky2009learning}, CIFAR-100 \cite{krizhevsky2009learning}, and MIO-TCD \cite{luo2018mio} datasets as well as a set of publicly available surveillance videos from static webcams that we collected and annotated.  More details about the data collection is given in the supplemental material.  In total we used 187,793 RGB training examples, resized to 64x64 pixels.

\textbf{Neural Architecture:}
We use a modified ResNext-50 network \cite{xie2017aggregated} as our backbone architecture. We modified the original model by adding an extra fully connected layer that maps the 2048-dimensional feature vector after the average pooling layer to a 128-dimensional layer. The 128-dimensional layer is then mapped by a final fully connected layer to an 8-dimensional output layer with sigmoid activations that represent the categories.  The extra fully connected layer gives us a 128-dimensional feature vector to represent appearance instead of the 2048-dimensional feature vector after RexNext-50's usual penultimate layer thus greatly improving memory efficiency.
To train our model, we utilize the Binary Cross Entropy as our loss function.  

Note that high-level features are usually distinctive even for unseen object classes despite the corresponding high-level attribute having low probabilities for all of the known object classes.  This allows our appearance model to handle object classes other than the eight that we train on.  See supplemental material for experiments on this.

\vspace{-5pt}
\subsubsection{Motion Model} To characterize the motion information for a given video volume, we train deep networks to estimate the following attributes directly from an RGB video volume: \textbf{(a)} histogram of optical flow  ($Y_{ang}$), \textbf{(b)} a vector of the average speed of pixels in each direction of motion ($Y_{speed}$), \textbf{(c)} background classifier ($Y_{bkg.cls}$) and \textbf{(d)} percentage of stationary pixels ($Y_{bkg.pix}$). The histogram of optical flow consists of 12 bins each of which stores the fraction of pixels in the video volume that are estimated to be moving in one of the 30 degree directions of motion.  The average speed vector consists of the average speed (in pixels per frame) of all pixels falling in each of the 12 histogram of flow bins.  The background classifier classifies whether the video volume contains motion or not.  The percentage of stationary pixels in a video volume gives the rough size of the moving objects in a video volume.

\textbf{Motion Training Data:}
We use the set of surveillance videos mentioned above to learn motion attributes in a self-supervised way.
For each video, we sample video volumes from regions with significant 'motion' as well as very little motion ('background'). We identify these regions by computing their pixelwise optical flow fields using the TV-L1 method \cite{zach2007duality}, which is also used to automatically generate ground-truth motion attributes.  (Note that optical flow is only used to create ground truth for training our motion models.  It is not used in later stages.)  
In total we obtain $283,486$ `background' video volumes and $2,551,376$ `motion' video volumes. We use $90\%$ of these for training our models and the remainder for validation.

The ground truth motion attributes for each training video volume are computed from the corresponding pixelwise flow fields as follows: 
We represent the $Y_{bkg.cls}$ attribute as a single binary variable denoting if a video volume is 'background'($Y_{bkg.cls} = 1$) or not ($Y_{bkg.cls} = 0$). 
The ground truth for $Y_{ang}$ and $Y_{bkg}$ are computed by first computing a 13-bin normalized histogram, wherein the first $12$ bins represent the number of pixels with flow orientation in the ranges $[i*\pi/6 : (i+1)*\pi/6 )$ with $i\in [0,11]$ , while the last bin denotes the number of pixels with flow magnitude below threshold.  The histogram is then normalized by the total number of pixels. 
The first 12 bins of this histogram are used as the ground truth for $Y_{ang}$ and the $13^{th}$ bin is the ground truth for $Y_{bkg}$.   Finally, we represent $Y_{mag}$ as a $12$-dimensional vector denoting the average flow magnitude for pixels in each of the 12 flow orientation ranges.

\textbf{Learning Task and Neural Architecture:} We treat each motion attribute independently and train separate models respectively. For each attribute prediction task, we use the same backbone architecture design, but train each model using different objective functions. Our model is a stack of 3D convolutions (3DConv) with batch normalization (BN) and ReLU. In total we have 3 layers of [3Dconv-BN-ReLU] followed by a fully connected layer. We provide additional details in the supplemental material.

We formulate $Y_{bkg.cls}$ attribute prediction as a standard binary classification task and train the model using cross-entropy loss function. For $Y_{bkg}$ and $Y_{mag}$ attribute prediction, we treat the learning task as a regression problem and train the model using mean squared error loss. And finally, for training the model to predict $Y_{ang}$ attribute, we utilize KL Divergence loss. For all the tasks, we construct a simple light-weight CNN. The detailed configuration of our 3D CNN architecture is presented in the supplemental material.
\subsection{Model Building}

Once trained, the attribute deep nets are used to build a model of any scene given the nominal video. As illustrated in Figure \ref{fig:modelbuilding}, to process each nominal video, we slide a spatio-temporal window of dimension $[h \times w \times t]$ with spatial stride $(h/2,w/2)$ and temporal stride of $t$ to construct video volumes.  In the experiments, we select $h=w$ and choose $h$ to be roughly the height in pixels of a person in a particular dataset.  For each RGB video volume, we extract its features using the previously trained appearance net and four motion nets. To get a single appearance feature vector for a video volume, the feature vectors computed by the appearance network for each frame of the video volume are averaged.
We concatenate the feature vectors from the penultimate layers of the appearance, angle, speed and background pixel nets along with the binary output of the background classifier net to create a combined feature vector.
We use $F$ to denote a combined feature vector and $app, ang, mag$, and $bkg$ to denote the appearance, angle, magnitude and background pixel fraction feature vectors, each of size $1\times 128$. Finally, $cls$ denotes the binary background classification of size $1 \times 1$.  $F$ is of size $1 \times 513$.

After computing features, we use the exemplar selection approach of \cite{jones2016exemplar, ramachandra2020learning} to create a region-specific compact model of the nominal data. For each region, we use the following greedy exemplar selection algorithm:

\vspace{-4pt}
\begin{enumerate}
\itemsep-0em 
    \item Add the first feature vector to the exemplar set.
    \item For each subsequent feature vector, compute its distance to each feature vector in the exemplar set and add it to the exemplar set only if all distances are above a threshold, $th$.
\end{enumerate}
\vspace{-4pt}

To compute the distance between two feature vectors $F_1 = [app_1; ang_1; mag_1; bkg_1; cls_1]$ and $F_2 = [app_2; ang_2; mag_2; bkg_2; cls_2]$ we use $L_2$ distances between corresponding components normalized by a constant to make the maximum distance for each component approximately 1.  When a video volume does not contain motion (as determined by the background classification, $cls$), the motion component vectors are set to 0.    The distance function can be written as follows:
\begin{equation}
d_{A}(F_1, F_2) = \lVert A_1 - A_2 \rVert_{2}\\
\end{equation}
where $A \in \{app, ang, mag, bkg\}$,
\begin{equation}
\resizebox{0.8\hsize}{!}{$
    d(F_1, F_2) = \frac{d_{app}}{Z_{app}} + \frac{d_{ang}}{Z_{ang}} + \frac{d_{bkg}}{Z_{bkg}} + \frac{d_{mag}}{Z_{mag}}
    $}.
    \label{eq:dist}
\end{equation}
The normalization factors, $Z_{app}$, $Z_{ang}$, $Z_{mag}$ and $Z_{bkg}$ are computed once by finding the max $L_2$ distances between a large set of feature vector components computed from a validation set (UCSD Ped1 and Ped2 in our experiments). 

One big advantage of the exemplar learning approach is that updating the exemplar set in a streaming fashion is possible. This makes the approach scalable and adaptable to environmental changes over time.

\subsection{Anomaly Detection}
At test time, we process each test video in the same way (by sliding a $[h \times w \times t]$ spatio-temporal window with spatial stride $(h/2, w/2)$ and temporal stride of $t$) to generate video volumes. For each video volume, we compute the combined feature vector as before using the pre-trained nets. Each combined feature vector is compared with every exemplar for the corresponding region using the distance function in Equation \ref{eq:dist}. The anomaly score for the given test video volume is the minimum distance over the set of all exemplars from the same spatial region.  A pixelwise anomaly score map is maintained by assigning the anomaly score to all pixels corresponding to every frame of the video volume. If a pixel has already been assigned an anomaly score (because of partially overlapping video volumes), then the maximum of the previous score and the current score is assigned.  Figure \ref{fig:modelbuilding} shows our anomaly detection pipeline.
\label{sec:method}

\subsection{A Note on Computational Efficiency}

For both model building and anomaly detection, most of the time is spent computing feature vectors (forward passes of 5 networks).  This is greatly sped up by testing whether a video volume is the same as the previous video volume in time.  If there is no change then the anomaly score for the new video volume should be the same as the one before it and no computation of feature vectors is needed.  This allows our method to run at 20 to 100 fps (dataset dependent).  Details are in the supplemental material.


\section{Experiments}
\subsection{Datasets and Evaluation Criteria}
We experiment on five benchmark datasets: UCSD Ped1 and Ped2 \cite{mahadevan2010anomaly}, CUHK  Avenue \cite{lu2013abnormal}, Street Scene \cite{ramachandra2020street} and ShanghaiTech \cite{luo2017revisit}. We use UCSD Ped1 and Ped2 with modified ground truth for parameter tuning and CUHK Avenue, Street Scene and ShanghaiTech for evaluation.
\\
\textbf{UCSD Ped1 \& Ped2:}  UCSD Ped1 dataset contains 34 training videos and 36 test videos while UCSD Ped2 dataset contains 16 training videos and 12 test videos. Anomalies consists of bikers, skaters and cars in a pedestrian area.
\\
\textbf{CUHK Avenue:} The Avenue \cite{lu2013abnormal} dataset contains 16 training videos with normal activity and 21 test videos. Examples of abnormal events in Avenue are related to people running, throwing objects or walking in wrong direction.
\\
\textbf{Street Scene:} The Street Scene \cite{ramachandra2020street} dataset contains 46 training videos defining the normal events and 35 test videos. Prominent examples of anomalies include jaywalking, loitering and bikes or cars driving outside their lanes.
\\
\textbf{ShanghaiTech:} The ShanghaiTech \cite{luo2017revisit} dataset is a multi-scene benchmark for video anomaly detection. It consists of 330 training and 107 test videos. Major categories of anomalies include people fighting, stealing, chasing, jumping, and riding bikes or skating in pedestrian zones.

While our primary focus is on single scene video anomaly detection task, we consider the ShanghaiTech dataset only to highlight the ease of usability and robustness of our method to multi-scene benchmarks. Our method is applied to ShanghaiTech without modification even though the location-dependent aspect of our model is not necessary for a multi-scene dataset.  Improvements in accuracy are likely if we specialize our model to multi-scene datasets.

\textbf{Evaluation Criteria.} We use the Region-Based Detection Criterion (RBDC) and the Track-Based Detection Criterion (TBDC) as proposed in \cite{ramachandra2020street} for quantitative evaluation of our framework.  These criteria correctly measure the accuracy of spatially and temporally localizing anomalous regions (RBDC) and anomalous tracks (TBDC) versus false positive detections per frame.  We report the area under the curve (AUC) for false positive rates per frame from 0 to 1 for each of these criteria. As pointed out in \cite{ramachandra2020street}, frame-AUC \cite{mahadevan2010anomaly} is not an appropriate evaluation  metric for video anomaly detection methods that spatially localize anomalies. However, we report frame-AUC scores of our method for completeness and comparison with other older methods. We also do not use the pixel-level criterion \cite{mahadevan2010anomaly} because of its serious flaws as mentioned in \cite{ramachandra2020street}. 

\subsection{Implementation}

\textbf{Feature Learning.} To train our \textbf{appearance model}, we use SGD with a 0.001 learning rate and 0.9 momentum and train for 50 epochs. The model with lowest classification error on the validation set is selected.  For \textbf{motion models} we optimize with AdamW~\cite{loshchilov2018decoupled} with a 0.001 learning rate and train for 30 epochs. We select the best model for each attribute using the validation set.
\\
\textbf{Video volume parameters.} We define the dimensions $(w,h)$ of a video volume for each dataset so that $h$ is roughly the height of a person in pixels and $w=h$. Specifically, for Ped1, Ped2, Avenue, Street Scene and ShanghaiTech, our region dimensions are $(32,32)$, $(32,32)$, $(128, 128)$, $(64,64)$ and $(100,100)$ respectively. Zero-padding was used for edge regions as needed. The number of frames in a video volume, t, is 10 for all datasets.

\begin{table}[tb]
  \centering
  \resizebox{\hsize}{!}{
  \setlength{\tabcolsep}{8pt}
  \begin{tabular}{|p{0.5cm}|c|c|c|c|c|c|}
    \hline
    \multirow{2}{*}{Th} & \multicolumn{3}{c|}{UCSD Ped1} & \multicolumn{3}{c|}{UCSD Ped2}\\
    \cline{2-7}
    & RBDC & TBDC & NUM & RBDC & TBDC & NUM\\
    \hline
    \hline
    3 & 36.866 & 77.83 & 288 & 64.808 & 89.13 & 350 \\ 
    2.5 & 49.36 & 89.43 & 424 & 78.813 & 93.716 & 761 \\ 
    2 & 57.524 & 89.6 & 944 &  84.66 & 95.97 & 1339 \\ 
    \bf{1.5} & \bf{61.65} & \bf{88.9} & \bf{4201} & \bf{87.44} & \bf{95.08} & \bf{4470} \\ 
    1 & 61.496 & 87.54 & 19926 & 87.408 & 95.776 & 19138 \\ 
    0.5 & 61.435 & 87.72 & 49113 & 87.195 & 95.12 & 34862 \\
    0.25 & 61.49 & 87.81 & 57636 & 87.199 & 95.127 & 45795 \\ 
    \hline
    
  \end{tabular}
  }
  \caption{RBDC and TBDC scores (in \%) of our method for different thresholds ($th$) on UCSD Ped1 and Ped2. NUM denotes the total number of exemplars across all regions.}
  \label{ucsd}
  \vspace{-10pt}
\end{table}

\textbf{Parameter Tuning.} To set a threshold $th$ for exemplar selection without fitting to test data, a validation data set is needed. We chose Ped1 and Ped2 for this purpose, both because these data sets are  performance-saturated, and because previous works~\cite{ramachandra2020learning} have identified inconsistencies in their ground truth. Specifically,
ground-truth annotations of Ped1 and Ped2 do not label every location-specific anomaly.
To rectify this, we augment the existing ground truth annotations to include all anomalies consistent with Definition 1.  This is justified because we are using Ped1 and Ped2 to set our hyperparameters and not to compare against previous methods.  Table~\ref{ucsd} shows region-based and track-based AUC for different values of the threshold $th$ used for exemplar selection for both Ped1 and Ped2.  We see that the accuracy of our method is robust to large variations of $th$.  However, larger values of $th$ lead to smaller numbers of exemplars and thus smaller models of the nominal video which is desirable. We select $th = 1.5$ as a good trade-off between accuracy and model-size.  We use this value on all datasets in our experiments. For Ped2, the average number of exemplars selected per region is about 13 ($\approx 0.5\%$ of the total number of video volumes in the nominal video).
Exemplar selection typically finds tens to sometimes low hundreds of exemplars (for Street Scene) for regions with lots of activity.  Regions with very little activity typically have only 1 or 2 exemplars.  This leads to very compact models of the nominal video.

\subsection{Quantitative Results}
Tables \ref{t1} and \ref{t2} compare our method to other top methods on Avenue, ShanghaiTech and Street Scene.  On Avenue, we improve over all previous methods for the region-based detection criterion (RBDC) and are second best for the track-based detection criterion (TBDC).  On ShanghaiTech, we improve over the next best method for both RBDC and TBDC by significant margins.  For the frame-level criterion which does not measure spatial localization we are in the middle of the pack compared to other recent methods for both Avenue and ShanghaiTech. On the  difficult Street Scene dataset (Table \ref{t2}), we improve the previous state of the art for both RBDC and TBDC, the latter by more than 11\%.  The good results across five different datasets (including Ped1 and Ped2) show the generality of the high-level features that we use in our models.

\begin{table}[tb]
  \centering
  \resizebox{\hsize}{!}{
\setlength{\tabcolsep}{8pt}
  \begin{tabular}{|p{3cm}|c|c|c|c|c|c|}
    \hline
    \multirow{2}{*}{Method} & \multicolumn{3}{c|}{Avenue} & \multicolumn{3}{c|}{ShanghaiTech}\\
    \cline{2-7}
    & RBDC & TBDC & Frame & RBDC & TBDC & Frame\\
    \hline
    \hline
    Ionescu \textit{et al}.\cite{ionescu2019object} & 15.77 & 27.07 & 87.4 & 20.65 & 44.54 & 78.7 \\
     Ramachandra \textit{et al.} \cite{ramachandra2020street}&  35.80 & 80.90 & 72.0 & - & - & -\\
     Ramachandra \textit{et al.} \cite{ramachandra2020learning}  & 41.20 & 78.60 & 87.2 & - & - & -\\
     Georgescu \textit{et al.} \cite{georgescu2021anomaly}  & 57.00 & 58.30 & 91.5 & 42.80 & 83.90 & \Red{\bf{90.02}} \\
     Liu \textit{et al.} \cite{liu2018future} &  19.59 & 56.01 & 85.1 & 17.03 & 54.23 & 72.8 \\
     Liu \textit{et al.} \cite{liu2021hybrid} &  41.05 & 86.18 & 89.9 & 44.41 & 83.86 & 74.2 \\
     Georgescu \textit{et al.} \cite{georgescu2021background} &  65.05 & 66.85 & \Blue{\bf{92.3}} & 41.34 & 78.79 & 82.7\\
     Liu \textit{et al.}\cite{liu2018future} $+$ Ristea \textit{et al.} \cite{ristea2021self} &  20.13 & 62.30 & 87.3 & 18.51 & 60.22 & 74.5 \\
     Liu \textit{et al.}\cite{liu2021hybrid} $+$ Ristea \textit{et al.} \cite{ristea2021self} &  62.27 & \Red{\bf{89.28}} & 90.9 & \Blue{\bf{45.45}} & \Blue{\bf{84.50}} & 75.5\\
     Georgescu \textit{et al.}\cite{georgescu2021background} $+$ Ristea \textit{et al.} \cite{ristea2021self} &  \Blue{\bf{65.99}} & 64.91 & \Red{\bf{92.9}} & 40.55 & 83.46 & \Blue{\bf{83.6}}\\
     Our Method &  \Red{\bf{68.2}} &  \Blue{\bf{87.56}} & 86.02 & \Red{\bf{59.21}} & \Red{\bf{89.44}} & 76.63\\
    
    \hline
    
  \end{tabular}
  }
  \caption{RRBDC, TBDC and Frame AUC scores (in \%) of various state-of-the-art methods on Avenue and ShanghaiTech datasets. The top score for each metric is highlighted in \Red{\bf{red}}, while the second best score is in \Blue{\bf{blue}}. }
  \label{t1}
  \vspace{-10pt}
\end{table}

\begin{table}[tb]
\setlength{\tabcolsep}{8pt}
\centering
\resizebox{0.7\hsize}{!}{
\begin{tabular}{| c | c | c | }
    \hline
    \bf{Methods} & RBDC & TBDC\\
    \hline
    \hline 
     Auto-encoder \cite{hasan2016learning} & 0.29 & 2.0 \\
     Dictionary method \cite{lu2013abnormal} & 1.6 & 10.0 \\
     Flow baseline \cite{ramachandra2020street} & 11.0 & 52.0 \\
     FG Baseline \cite{ramachandra2020street} & \Blue{\bf{21.0}} & \Blue{\bf{53.0}} \\
     Our Method &  \Red{\bf{24.26}} & \Red{\bf{64.5}}  \\
     \hline
\end{tabular}
}
\caption{RBDC and TBDC AUC scores (in \%) of various baseline methods on Street Scene dataset. The top score for each metric is highlighted in \Red{\bf{red}}, while the second best score is highlighted in \Blue{\bf{blue}}. }
\label{t2}
\vspace{-5pt}
\end{table}

\subsection{Qualitative Results: Explainability}
\label{sec:explainability}

\begin{figure*}[tb]
    \centering
    \includegraphics[width=0.9\linewidth]{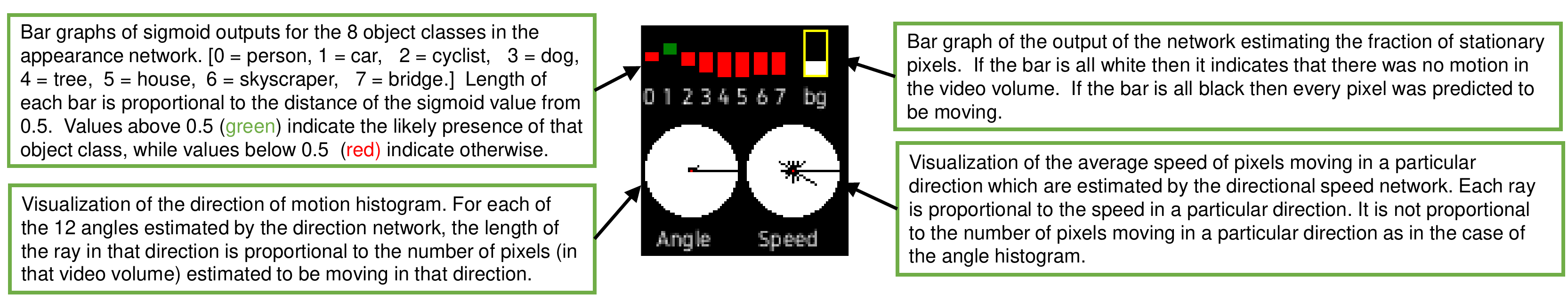}
    \caption{Explanation of our ``instrument panel'' showing the estimated attributes for a video volume. The interpretation of this visualization would be (roughly) a car (class 1) taking up most of the video volume, moving right at a high speed.}
    \label{fig:viz_explanation}
\vspace{-10pt}
\end{figure*}

\begin{figure}[tb]
    \centering
    \includegraphics[width=0.48\textwidth]{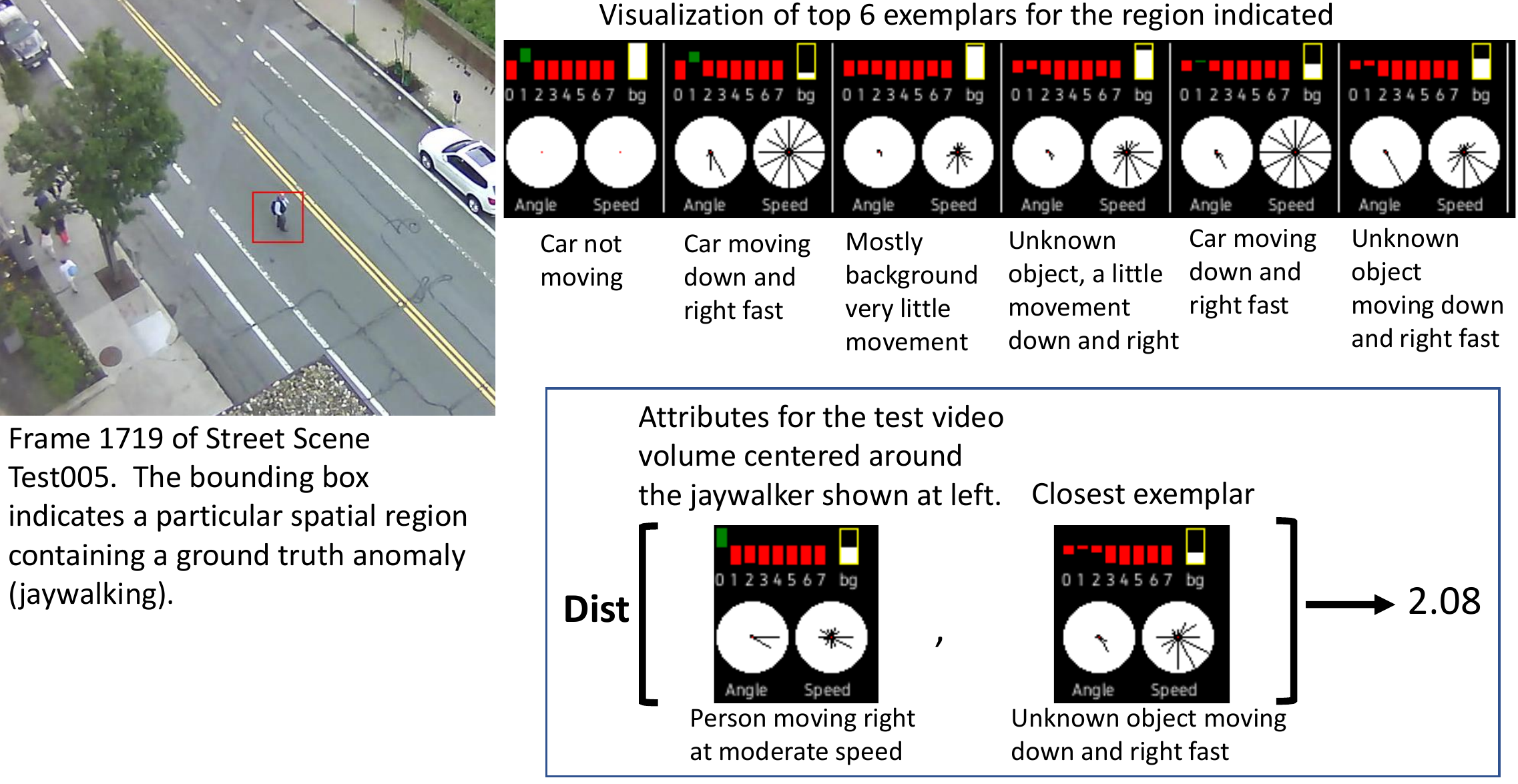}
    \caption{Visualization of the learned exemplars for a region of Street Scene and visualization of a test video volume explaining why it was detected as an anomaly.}
    \label{fig:visualization_jaywalk}
\vspace{-8pt}
\end{figure}

One of the big advantages of our method in addition to its accuracy is that it allows intuitive explanations of what the model has learned and why it labels a particular test video volume as anomalous or not.  To visualize the feature vector representing a video volume, the appearance and motion components of the combined feature vector are mapped using the last fully connected layer of the respective network to the high-level appearance and motion attributes.  We can then visualize these attributes as illustrated in Figure \ref{fig:viz_explanation}.

As an illustration of our model's explainability, the top of Figure \ref{fig:visualization_jaywalk} visualizes the exemplars learned from the nominal video for a spatial region in the middle of the street.
Cars travel down and to the right in this lane of the street.  The top six exemplars learned show mainly cars (or unknown objects, since video volumes containing only parts of cars are often not classified as cars) traveling down and to the right, as expected.  There are also exemplars for stationary background, as well as stationary cars (since occasionally traffic stops on this part of the street).  Thus, our learned model is understandable and consistent with what one expects.
Furthermore, for a video volume containing a person jaywalking, the visualization in the bottom, left of the figure shows that our networks correctly identify it as containing a person walking mainly to the right at moderate speed.  The closest exemplar to this test volume is an unknown object moving down and to the right  which yields a high anomaly score of 2.08.  (A threshold of 1.8 yields high detection rates with low false positive rates across all the datasets.)  Thus, the explanation of this anomaly is that there is an unusual object (person) walking in an unusual direction.

\begin{figure}[tb]
    \centering
    \includegraphics[width=0.48\textwidth]{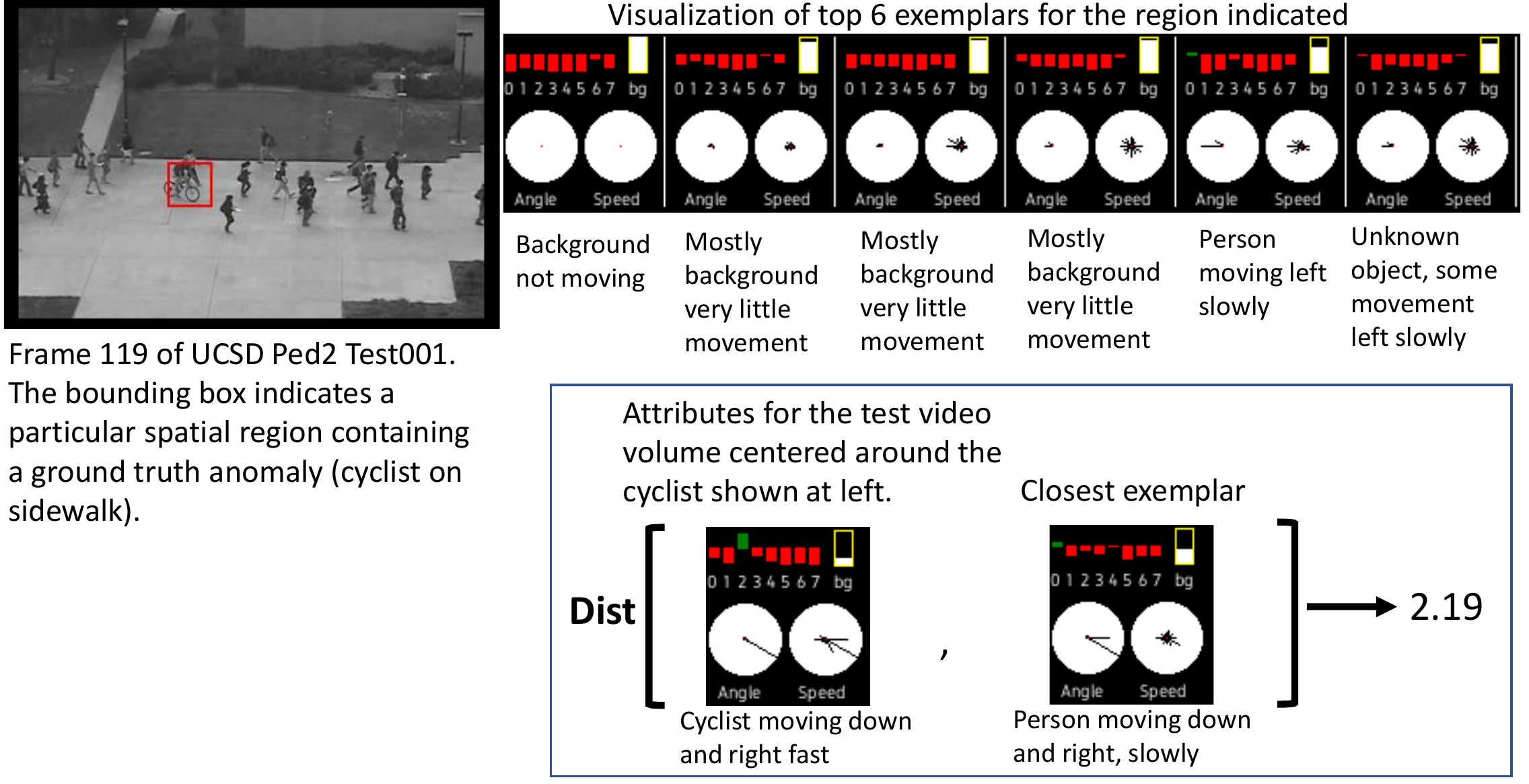}
    \caption{Visualization example for a region of UCSD Ped2 showing an explanation of the anomaly.}
    \label{fig:visualization_ped2}
   \vspace{-10pt}
\end{figure}

Another example is shown for Ped2 in Figure \ref{fig:visualization_ped2}.  Here we analyze a region on the sidewalk.  The exemplars learned for this region (shown at the top of Figure \ref{fig:visualization_ped2}) are mainly background with very little movement or people moving mainly left or right at slow speeds. Some video volumes containing only parts of people are not classified as people which leads to exemplars of unknown objects moving left or right.  Overall, these exemplars are again what we would expect for this region.  For the test frame shown, a cyclist is riding on the sidewalk.  The visualization of the video volume centered on that frame at that spatial region shows that it was classified as a cyclist traveling down and right at a high speed.  These high-level attributes differ from the nearest exemplar in terms of its object class and speed and therefore leads to a high anomaly score.

\begin{figure}[tb]
    \centering
    \includegraphics[width=0.48\textwidth]{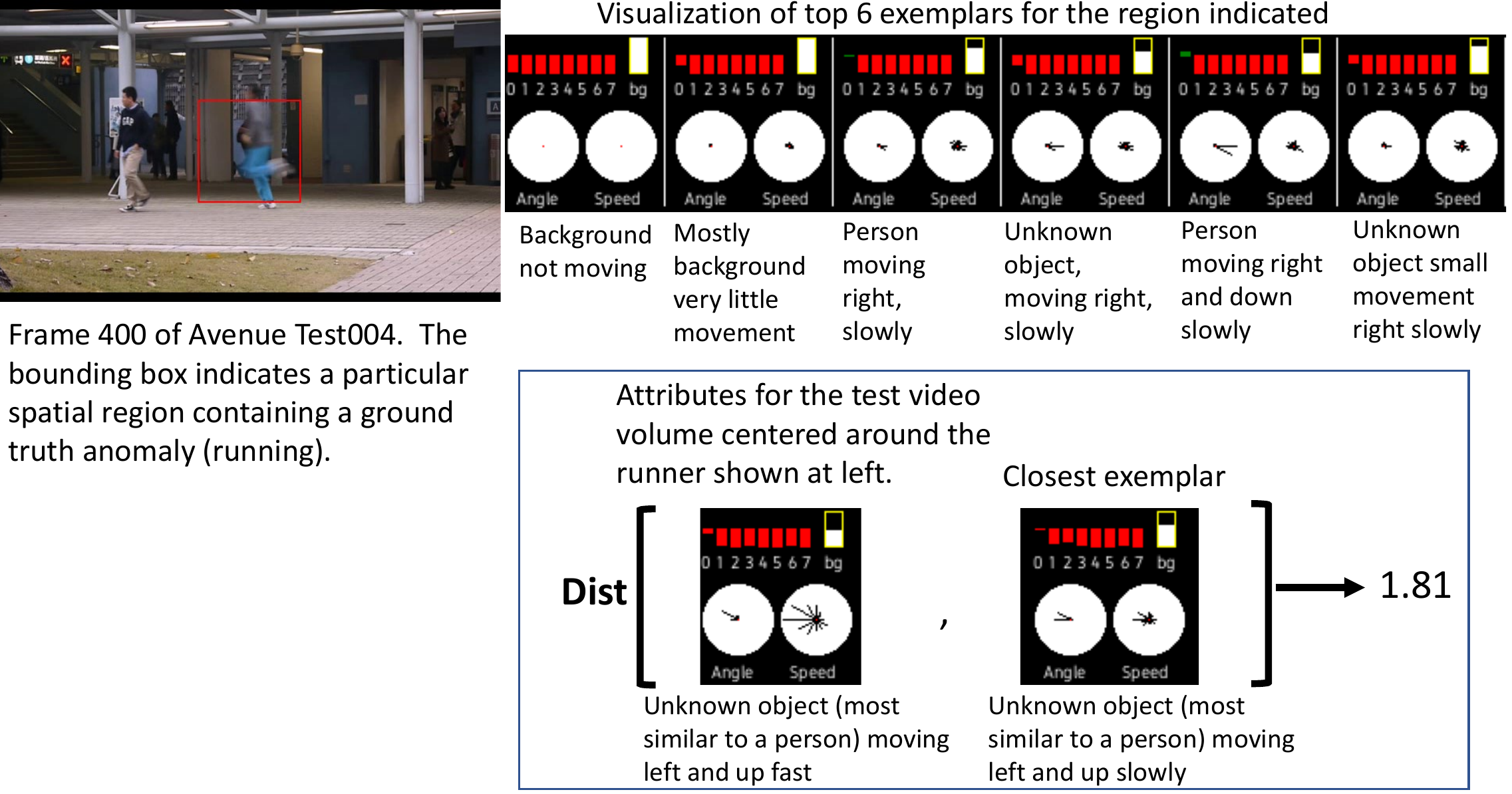}
    \caption{Visualization example for a region of CUHK Avenue showing an explanation of the anomaly.}
    \label{fig:visualization_avenue}
\end{figure}

As a final illustration (we show more in the supplemental material), we look at an example from CUHK Avenue in Figure \ref{fig:visualization_avenue}. 
The top six exemplars for the region highlighted at the left of the figure show that the model has learned that this region contains either background with very little movement or else people or unknown objects moving mainly left or right slowly.  For the anomalous test video volume shown containing a person running to the left, the high-level features estimate an unknown object moving left at high speed.  Even though the object recognizer did not correctly predict that the video volume contains a person, the person class is the most likely out of the eight classes.  The nearest exemplar is an unknown object (closest to a person class) moving slowly to the left.  The main difference between the test video volume and the closest exemplar is the unusual speed which correctly explains this anomaly.

\label{sec:experiments}


\subsection{Ablation Study}
\label{sec:ablation}

We perform an ablation study on all the benchmarks to evaluate the benefit of each attribute in detecting anomalies. We consider features from each model separately, combining features from only motion models and finally our full model which uses all features. We accordingly change the distance function to compare features of two video volumes so that it only uses the provided features. 
We present our results in Table \ref{ablation1}. We see that different attributes can be important for different types of scenarios. However, we get best results across all the benchmarks only when we combine all the motion and appearance features.
This highlights the importance of modeling both appearance and different components of motion, especially to be able to predict anomalies under wide variety of scenarios.

\begin{table}[t]
\setlength{\tabcolsep}{1.5pt}
\centering
\begin{tabular}{| c | c | c | c | c |}
    \hline
    Attributes & Ped1 & Ped2 & Avenue & Street Scene\\
    \hline
    \hline 
     App & 29.6 / 64.8 & 77.7 / 92.5 & 75.6 / 67.4 & 1.1 / 4.8\\
     \hline
     Motion & 59.2 / 83.7 & 81.6 / 93.3 & 69.0 / 89.0 & 22.6 / 64.1\\
     \hline
     Angle &  40.9 / 70.6  &  70.8 / 88.1  &  66.4 / 89.3  &  24.5 / 65.9 \\
     \hline
     Mag &  60.6 / 90.1  &  70.1 / 88.5  &  59.5 / 91.1  & 16.6 / 50.8 \\
     \hline
     Bkg &  45.5 / 86.4  &  71.7 / 95.2  & 49.9 / 81.5  &  11.9 / 49.8 \\
     \hline
     App+Mot & 61.7 / 88.9 & 87.4 / 97.1 & 69.6 / 86.7 & 23.9 / 65.3\\
     \hline
\end{tabular}
\caption{RBDC / TBDC AUC scores (in \%) of our method when using only appearance, only motion (using angle, magnitude and background pixel predictions combined), each motion component separately and all the features}
\label{ablation1}

\end{table}

\begin{table}[t]
\setlength{\tabcolsep}{12pt}
\centering
\begin{tabular}{| c | c | c |}
    \hline
    Attributes & Ped1 & Ped2 \\
    \hline
    \hline 
     ImageNet & 25.148 / 44.63 & 64.67 / 83.17 \\
     \hline
     Ours & 29.6 / 64.8 & 77.7 / 92.5 \\
     \hline
\end{tabular}
\caption{RBDC / TBDC AUC scores (in \%) of our method when using our pre-trained model versus ImageNet pe-trained model as appearance feature extractor.}
\label{ablation2}
\vspace{-15pt}
\end{table}

We further perform an ablation study on UCSD Ped1 and Ped2 datasets to empirically evaluate the benefit of our appearance model to represent object features versus using ImageNet pre-trained features. For the ImageNet pre-trained model, we use the pre-classification layer output of ResNext-50 as features. 
We present our results in Table \ref{ablation2}. The superiority of our model is most likely due to the loss function used (binary cross-entropy) which allows an image patch to contain zero or multiple object classes.


\section{Discussion and Conclusions}
\label{sec:discuss}

We have presented a novel method for explainable video anomaly localization that has a number of desired properties.  Foremost, the method is accurate and general.  We have shown that it works very well on five different datasets and, in particular, achieves state-of-the-art results on CUHK Avenue, Street Scene and ShanghaiTech.  Setting it apart from almost all previous work, our model is understandable by humans and the decisions that our method makes are explainable.
Finally, because our method does not require a computationally expensive training phase on the nominal data, it is  easy to expand our model when new nominal data becomes available.


%% file: body_supplemental.tex
\section{Supplemental Material}

In this supplemental material, we will give more details on the high-level appearance and motion deep networks that we train including the training examples used and the networks' accuracies. We show more visualizations of the explainability of our method.  We discuss the speed of our method both for model-building and anomaly detection.  We also discuss the limitations of our approach.  Finally, we include a number of videos showing the anomaly detections of our method on videos from the Street Scene, Avenue, Ped1 and Ped2 datasets.

\subsection{Data generation}
\label{sec:data_gen}

\subsubsection{Webcam dataset}
Our key motivation is to learn features that can efficiently represent generic knowledge about outdoor environments.
To this end we collected surveillance videos from publicly available webcams. We collected 33 videos in total of length 3 minutes each on average.

\begin{figure*}[h]
    \centering
    \includegraphics[width=\linewidth]{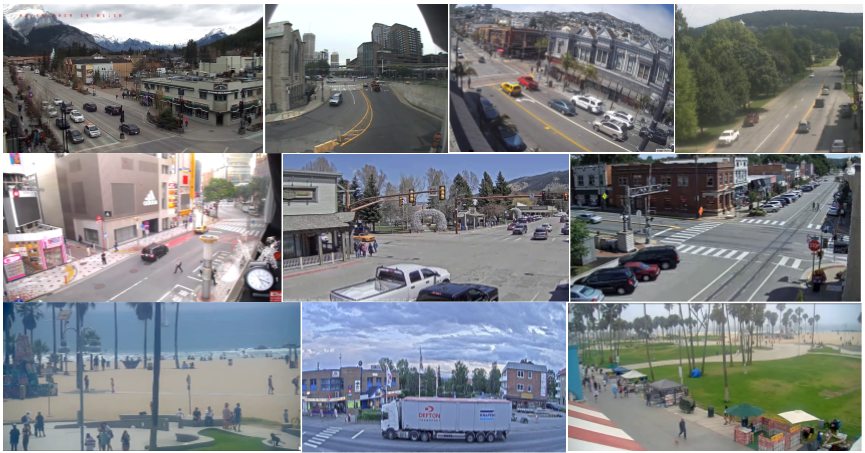}
    \caption{Selected frames from the webcam dataset of surveillance videos}
    \label{fig:webcam}
\end{figure*}

\subsubsection{Appearance Model}
As discussed in the main paper, we created our training dataset from multiple sources (CIFAR-10 \cite{krizhevsky2009learning}, CIFAR-100 \cite{krizhevsky2009learning}, and MIO-TCD \cite{luo2018mio} and webcam videos as discussed above).

 Many of our training examples, especially for the person, car and cyclist classes come from the webcam videos.  We manually annotated the videos with bounding boxes around the people, cars and cyclists.  In addition we added a subset of the car, pedestrian and cyclist examples from the MIO-TCD dataset \cite{luo2018mio}.  Finally, we used the car and dog examples from CIFAR-10 \cite{krizhevsky2009learning} and the tree, house, skyscraper and bridge classes from CIFAR-100 \cite{krizhevsky2009learning}. In total, we collected 116,799 images for training and 9,240 images held out for validation spread across the 8 classes, all resized to 64x64 pixels. 
 
 \begin{figure}[h]
    \centering
    \includegraphics[width=\linewidth]{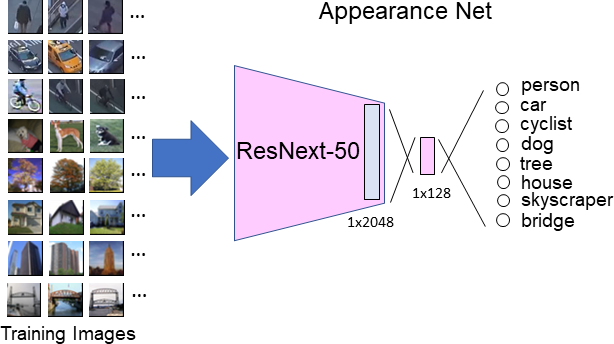}
    \caption{Sample images from each category used for training the appearance model as well as the basic architecture used for our appearance model.  The input to the network is a single 64x64 pixel color RGB image and the output is an independent probability (sigmoid function) for each of the eight output classes (not softmax).  Thus, there can be more than one object class recognized for a single input image.}
    \label{fig:appearance}
\end{figure}

 After initial training of a ResNext-50 network \cite{xie2017aggregated}, we scanned the resulting classifier across a set of 28 large images of scenes not containing any of the 8 object classes.  Any patches classified as one of the objects were collected to form a new set of hard negative examples.  This yielded an additional set of 62,336 background images which was added to the training set and a new object recognizer was trained from scratch.  Hard negative mining was done a second time to yield one more set of 8,658 background patches.  The total set of 187,793 images was used for a final training from scratch to yield the final classifier.

Figure \ref{fig:appearance} shows example 64x64 pixel training images for each class as well as the basic network architecture used for the appearance network.

\subsection{Motion Model}

As discussed in the main paper, we use RGB video volumes as input to our motion attribute networks and compute ground-truth attribute labels using optical flow. Every video volume in our dataset can be categorized as either `motion' or `background'. To create video volumes, we sequentially sample $N$ continuous frames from a video and the corresponding $N-1$ optical flow frames. We chose $N = 10$ frames to follow the same settings as our video anomaly detection pipeline. Using the flow frames and two fixed thresholds, we define  `regions with significant motion' and 'background regions'. The first threshold ($ th_{mot}$) is the maximum magnitude of a flow vector for it to be counted as a moving pixel.  The other threshold ($ th_{bkg}$) is the percentage of moving pixels required to say a video volume contains motion.  For our experiments we select $ th_{mot} = 1.0$ and $ th_{mot} = 99\%$.  We sample motion and background video volumes from their respective regions. We do this to improve the efficiency of selecting video volumes, as for most surveillance videos, only a small set of regions have some form of activity.

\begin{figure*}[h]
    \centering
    \includegraphics[width=0.8\linewidth]{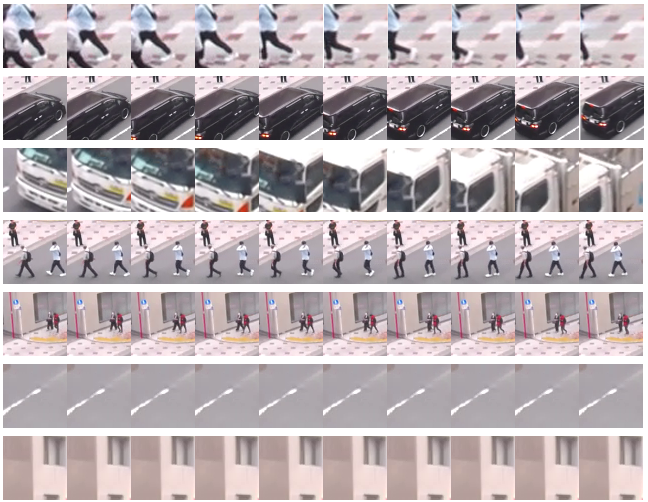}
    \caption{Examples of automatically generated video volumes for training motion attribute models. Rows 1-5 shows example video volumes from `motion' regions, while Rows 6-7 shows `background' video volumes. }
    \label{fig:motionvvols}
\end{figure*}
 
 We sample $2,551,376$ `motion' video volumes and $283,486$ 'background' video volumes. (Background samples are 10\% of the total samples.)
 After sampling, we resize all the video volumes to spatial dimension $[64 \times 64]$. Thus each video volume is of dimension $[64 \times 64 \times 3 \times N]$ ( $[h \times w \times c \times t]$ ) .
 We use $90\%$ of these for training our models and the remainder for validation.

\begin{figure}[h]
    \centering
    \includegraphics[width=\linewidth]{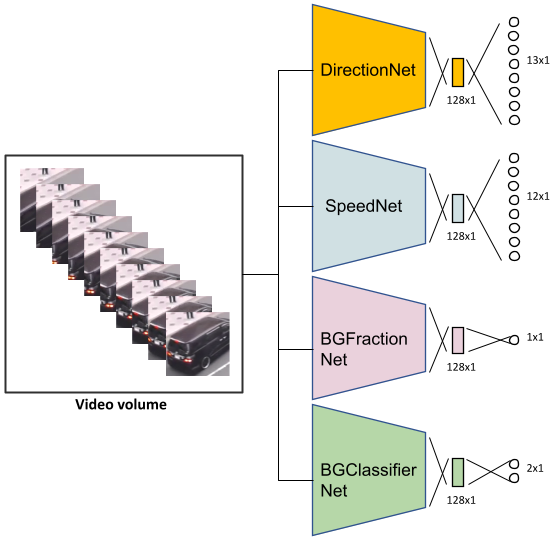}
    \caption{ Motion attribute models.}
    \label{fig:motionattribute}
\end{figure}

\subsection{Motion network Architecture}
\label{sec:motnet}

Our backbone convolutional neural network model for motion attribute learning is composed of three 3D convolution (conv) layers and three 3D max-pooling layers, followed by a fully-connected layer. Each conv layer is followed by a batch normalization layer and a ReLU activation. 

\begin{figure}[h]
    \centering
    \includegraphics[width=0.7\linewidth]{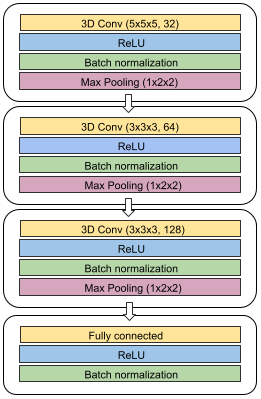}
    \caption{Backbone architecture for our motion model}
    \label{fig:motionmodel}
\end{figure}

The first 3D conv layers use filters of dimensions $5 \times 5 \times 5$, while the remaining two 3D conv layers use filters of dimensions $3 \times 3 \times 3$ each. The three 3D conv layers have 32, 64 and 128 filters respectively. We set the padding to “same” and stride to 1. We perform only spatial pooling for all three 3D max-pooling layers. The pooling size and the stride are both set to 2. We add a fully connected layer at the end to obtain a 128 dimensional feature vector. 

For all the motion attribute learning tasks, we train separate models. For each task-specific model, we use the same backbone architecture described above with an additional task specific prediction head. For angle prediction, we add a fully connected layer with 13 units. For magnitude and background pixel percentage prediction we add a fully connected layer with 12 units and 1 unit each. Finally, for background classifier, we add a fully connected layer with 1 output unit.

\subsection{Experiments on the Accuracy of Appearance and Motion Networks}

The results in the main paper on 5 different video anomaly detection datasets show that the features learned by our appearance and motion networks are very effective for detecting anomalies in video.  It is also interesting to analyze how accurate our networks are on the object recognition and motion attribute prediction tasks they are trained for.  Table \ref{tab:appearance_accuracy} shows correct detection and false positive rates for our appearance network on a held-out test set of 64x64 pixel images containing person, car, cyclist, dog, tree, house, skyscraper, bridge and background (none of the above) classes.  Overall, accuracy is quite good.  The cyclist class has the lowest accuracy due to the fact that for some views of cyclists, the bike is heavily occluded by the rider which can cause the cyclist to be classified as a person.  This also explains why the person class a somewhat higher false positive rate than other classes.

In Table \ref{tab:motion_accuracy} we show the error rates computed for each motion attribute network. Specifically, for 'Background classifier' (BGClassifierNet) we report the classification error percentage on the held-out validation set.  The table shows that the background classifier is correct over 98\% of the time (1.69\% error).  For the 'Background Fraction' (BGFractionNet) attribute model, we report the average L1 error.  This network outputs values between 0 and 1, so 0.053 average error is quite low.  In the case of the 'Angle' (DirectionNet) and 'Magnitude' (SpeedNet) attribute models, which output 12 values for the 12 different angle bins, we are interested in evaluating the average deviation of the predicted estimate to the ground-truth value over all possible angle bins. To this end we compute the mean of absolute difference for both the normalized angle histogram and the magnitude vector.  For DirectionNet, the output is a histogram so all values are between 0 and 1 and an average L1 error of 0.0184 shows good accuracy.  For SpeedNet, values do not have an upper bound but are typically between 0 and 10 pixels/frame.  An average L1 error of 0.331 shows low error.  Our results demonstrate that our models can accurately predict motion attributes for unseen videos with small errors.

Further improving network accuracy will lead to increases in video anomaly localization accuracy.

\begin{table}[tb]
\setlength{\tabcolsep}{8pt}
\centering
\resizebox{\hsize}{!}{
\begin{tabular}{| c | c | c |}
    \hline
    Class & Correct Detection Rate & False Positive Rate \\
    \hline
    \hline 
     person &  95.5\% & 3.6\% \\
     \hline
     car & 94.2\% & 1.8\% \\
     \hline
     cyclist & 77.6\% & 1.1\% \\
     \hline
     dog & 99.0\% & 0.3\% \\
     \hline
     tree & 99.0\% & 0.5\% \\
     \hline
     house & 89.0\% & 0.5\% \\
     \hline
     skyscraper & 97.0\% & 0.3\% \\
     \hline
     bridge & 97.0\% & 0.7\% \\
     \hline
\end{tabular}
}
\caption{Detection and false positive rates for our appearance network on a held-out test set of 64x64 pixel RGB images.}
\label{tab:appearance_accuracy}
\end{table}

\begin{table}[tb]
\setlength{\tabcolsep}{8pt}
\centering
\resizebox{0.5\hsize}{!}{
\begin{tabular}{| c | c |}
   \hline
    Attributes & Error Rate \\
    \hline
    \hline 
     BGClassifierNet &  1.69\% \\
     \hline
     BGFractionNet & 0.053 \\
     \hline
     DirectionNet & 0.0184 \\
     \hline
     SpeedNet &  0.331\\
     \hline
\end{tabular}
}
\caption{Error rates for our motion networks on a held-out test set of 10x64x64x3 video volumes.}
\label{tab:motion_accuracy}
\end{table}

\subsection{How well do appearance feature vectors for unknown classes cluster together?}

In the introduction we mention that video volumes containing unknown object classes do not cause a problem for our method because the appearance feature vectors (output by our appearance network) for different images of the same object class tend to have small distance.  This is the main advantage of using the network's embedding as our appearance feature as opposed to using the output class probabilities.  In order to back this claim up with data, we used a set of 1000 horse images and 1000 ship images from Cifar-10 which are very different object classes from the 8 classes our appearance network was trained on.  For each image, we computed its embedding using our appearance network and then computed separately the average $L_2$ distance between all horse images, between all ship images and between horse and ship images.  The average
$L_2$ distance between horse image embeddings was 11.1, the average distance between ship image embeddings was 18.3, and the average distance between horse versus ship embeddings was 22.0.  This shows that embeddings for images of the same class tend to be closer than embeddings for images of different classes.

Furthermore, we ran k-means clustering using two clusters on the horse and ship embeddings.  The two resulting clusters approximately separated the two object classes.  One cluster contained 91\% horse embeddings (and 9\% ship embeddings) and the other cluster contained 77\% ship embeddings and 23\% horse embeddings.  Again this shows that the embedding learned by our object recognizer does a good job of clustering unknown object classes.

\subsection{Computational Analysis}
We analyze computational speed of our method on the Ped2, Avenue and Street Scene datasets. For each dataset, we compute the processing speed for anomaly detection stage.  The running time for model building (exemplar selection) is almost identical to anomaly detection.  We compute the total time taken by adding the time taken to extract features from our high-level models and perform nearest neighbor matching.  The main computational bottleneck for our method is computing feature vectors, which requires evaluating 5 different neural networks, on every video volume.  A simple but effective method was used to speed this up.  The important insight is that the feature vector for a video volume should not change from one time step to the next if the pixels of the video volume have not changed.  If the feature vector does not change then the anomaly score will not change either.  So, for any video volume that is almost identical to the previous video volume in time, we do not need to compute its feature vector and the anomaly score for the previous video volume can simply be used for the new video volume.  We use normalized cross correlation to determine whether two video volumes are nearly identical.  Note that this speed-up does not prevent our method from detecting static anomalies (such as loiterers).

For each dataset, the size of the spatial regions and thus the number of regions differs since it is chosen depending on the approximate height of a person in the dataset. Furthermore, the size of frames in each dataset differs.  As a result, the computational speed differs for each dataset.

\begin{table}[H]
\setlength{\tabcolsep}{8pt}
\centering
\resizebox{0.7\hsize}{!}{
\begin{tabular}{| c | c |}
    \hline
    Dataset & Anomaly Detection \\
    \hline
    \hline 
     Ped2 &  32 fps \\
     \hline
     Avenue & 112 fps \\
     \hline
     Street Scene & 12 fps \\
     \hline
\end{tabular}
}
\caption{Computational speed for our pipeleine. We show speed for each stage in frames/second.}
\label{tab:our_speed}
\end{table}

We present our results in Table \ref{tab:our_speed}. For each dataset, we report results in frames per second. We used a single NVIDIA Quadro RTX 8000 GPU for feature extraction and Intel Xeon E5-2680 v4 @ 2.40GHz CPU for nearest neighbour computations. 
For the Avenue dataset with $640 \times 360$ resolution frames and a region-size of $128 \times 128$ resulting in 45 regions, the speed is relatively fast at over 6 frames/sec.  For Ped2 (with $360 \times 240$ frames and 345 spatial regions) and especially for Street Scene (with $1280 \times 720$ frames and 897 spatial regions) our method is under 1 frame/sec.

We also show in Table \ref{tab:other_speed} the running times for other published VAD methods.  These times are for the anomaly detection phase only.  (Note that different methods are benchmarked using different GPUs so the numbers are not directly comparable.)  For the model building phase, most other methods require training a deep network on the nominal video which makes those methods much slower than ours since ours requires no network training in the model building (exemplar learning) or anomaly detection stages.

\begin{table}[H]
\setlength{\tabcolsep}{8pt}
\centering
\resizebox{\hsize}{!}{
\begin{tabular}{| c | c | c |}
    \hline
    Method & Detection Speed & GPU type\\
    \hline
    \hline
     Ionescu et al \cite{ionescu2019object} & 11 fps & Titan XP \\
     \hline
     Georgescu et al \cite{georgescu2021anomaly} &  21 fps & GTX 1080Ti \\
     \hline
     Georgescu et al \cite{georgescu2021background} & 18 fps & GTX 3090 \\
     \hline
     Liu et al \cite{liu2018future} & 25 fps & GeForce TI-TAN \\
     \hline
     Liu et al \cite{liu2021hybrid} & 10 fps & RTX 3090 \\
     \hline
     Ours & 12 to 112 fps & Quadro RTX 8000 \\
     \hline
\end{tabular}
}
\caption{Computational speed for our pipeleine. We show speed for each stage in frames/second.}
\label{tab:other_speed}
\end{table}

\subsection{Example Result Frames}

\begin{figure}[thb]
    \centering
    \includegraphics[width=0.75\linewidth]{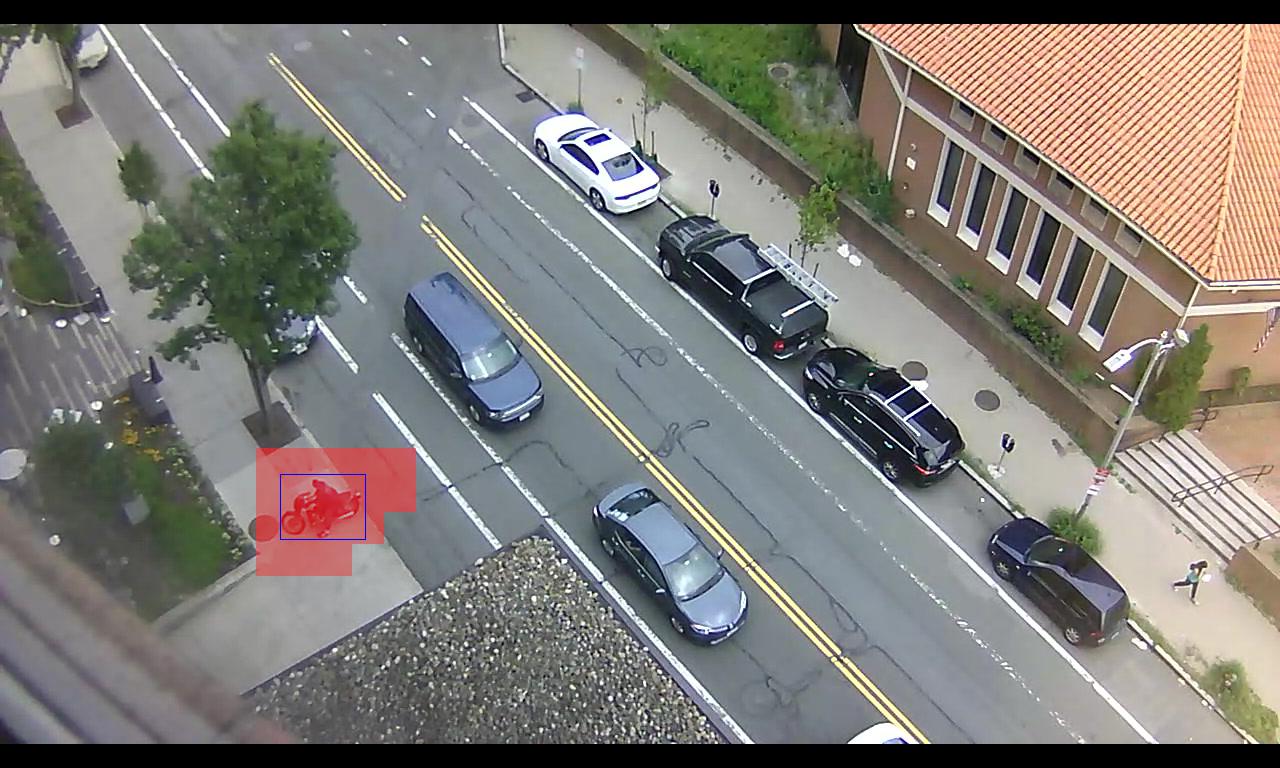}
    \caption{A test frame from Street Scene (Test031) showing the areas detected as anomalous by our method (shaded in red) and the ground truth bounding box in blue.}
    \label{fig:motorcycle}
\end{figure}

\begin{figure}[thb]
    \centering
    \includegraphics[width=0.75\linewidth]{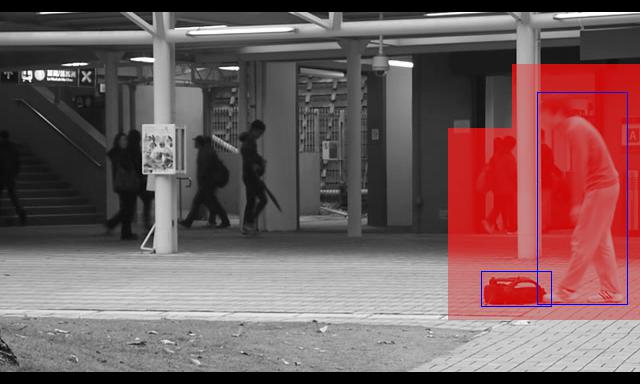}
    \caption{A test frame from CUHK Avenue (Test006) showing the areas detected as anomalous by our method (shaded in red) and the ground truth bounding boxes in blue.}
    \label{fig:avenuedetections}
\end{figure}

\begin{figure}[thb]
    \centering
    \includegraphics[width=0.7\linewidth]{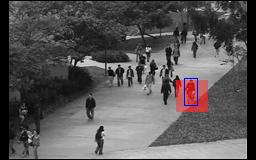}
    \caption{A test frame from UCSD Ped1 (Test006) showing the areas detected as anomalous by our method (shaded in red) and the ground truth bounding box in blue.}
    \label{fig:ped1detections}
\end{figure}

\begin{figure}[thb]
    \centering
    \includegraphics[width=0.7\linewidth]{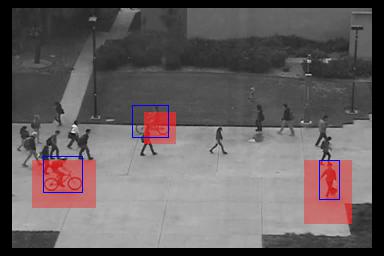}
    \caption{A test frame from UCSD Ped2 (Test006) showing the areas detected as anomalous by our method (shaded in red) and the ground truth bounding boxes in blue.}
    \label{fig:ped2detections}
\end{figure}

We show a few frames from Street Scene, CUHK Avenue, UCSD Ped1 and Ped2 with the areas detected as anomalous from our method shaded in red and the ground truth anomalies shown as blue bounding boxes in Figures \ref{fig:motorcycle} - \ref{fig:ped2detections}. 
We also include example results videos from each datasets in our supplementary material.

\subsection{Additional Visualizations of Results}

\begin{figure*}[tb]
    \centering
    \includegraphics[width=0.8\linewidth]{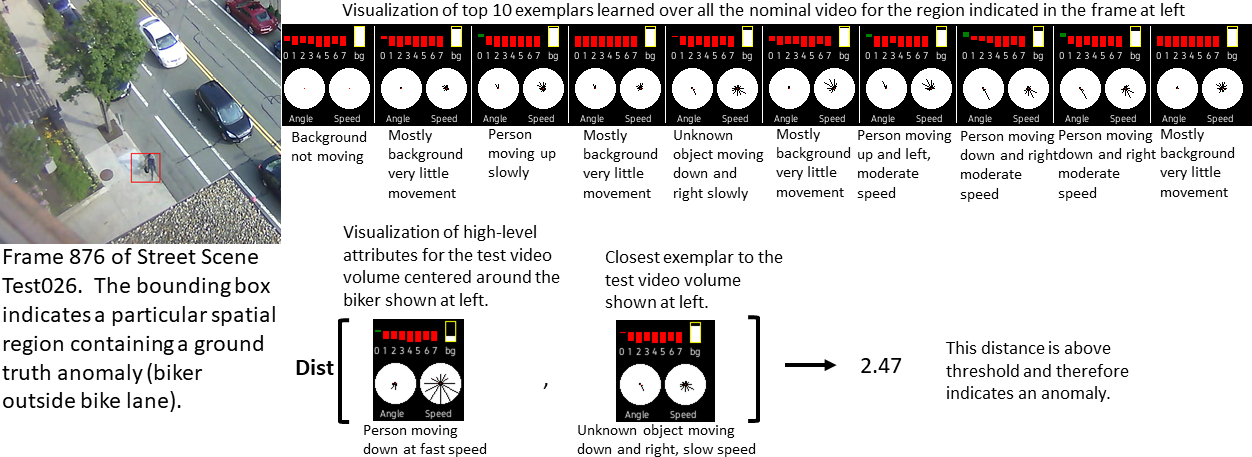}
    \caption{Visualization of the learned exemplars for a region of Street Scene and visualization of a test video volume explaining why it was detected as an anomaly.}
    \label{fig:visualization_biker}
\end{figure*}

Figure \ref{fig:visualization_biker} shows a visualization of the exemplars learned for a region of Street Scene on the edge of the sidewalk as well as a visualization of the high-level attributes estimated for a video volume in this region around an anomalous cyclist who is outside of the bike lanes (shown on the left side of the image).  The top ten exemplars for this region (along the top of the figure) show either background/unknown objects with little motion or people moving in the direction of the sidewalk at low speed, as expected.  The visualization of the high-level attributes for the video volume centered on the cyclist shown in the frame on the left, show that the video volume was estimated to contain a person moving downward at a fast speed.   Although the object class is incorrect (it should be class 2, cyclist), the direction and speed are still different from the exemplars learned for this region.  The closest exemplar (shown at the bottom right of the figure) is estimated to contain an unknown object (although person is the most likely class) moving down and to the right at a slow speed.  The distance between the test feature vector and the closest exemplar feature vector is 2.47 which is high and indicates an anomaly.

\begin{figure*}[tbh]
    \centering
    \includegraphics[width=0.8\linewidth]{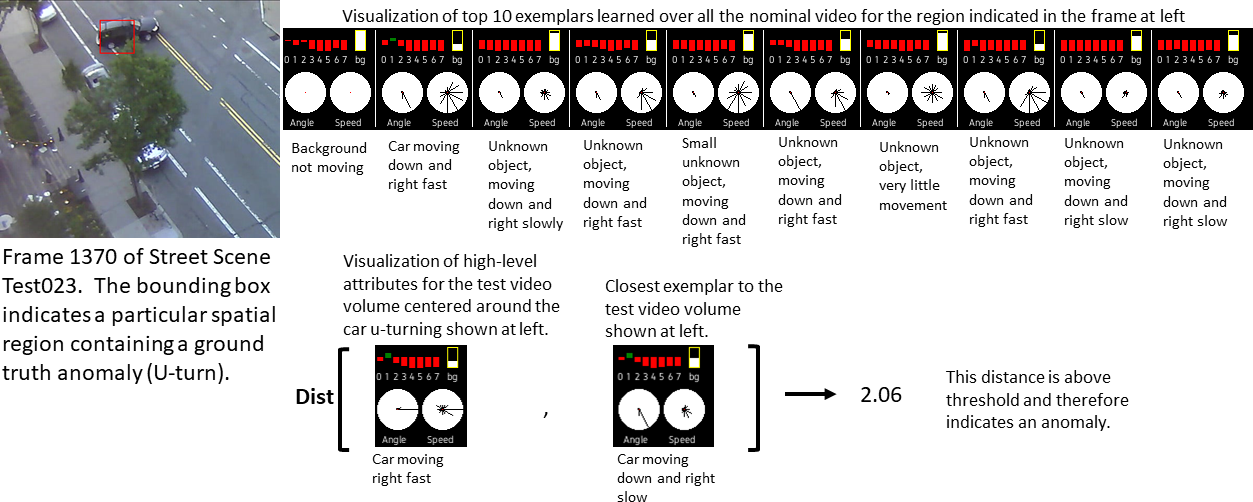}
    \caption{Visualization of the learned exemplars for a region of Street Scene and visualization of a test video volume explaining why it was detected as an anomaly.}
    \label{fig:visualization_uturn}
\end{figure*}

Figure \ref{fig:visualization_uturn} shows a region of Street Scene on the street.  As expected, the visualization of the top ten exemplars shows either background with little or no movement or cars/unknown objects moving mainly down and right (the direction of the street) at various speeds.  The attributes of a video volume centered around a car that is making a u-turn is visualized at the bottom, left of the figure.  It shows a car moving right at a fast speed.  The nearest exemplar is a car moving down and right at a slow speed.  The exemplar-based model does not have any examples of cars moving in this direction from the nominal data.  Therefore, the test video volume has a high anomaly score and is detected as anomalous.

\begin{figure*}[tbh]
    \centering
    \includegraphics[width=0.8\linewidth]{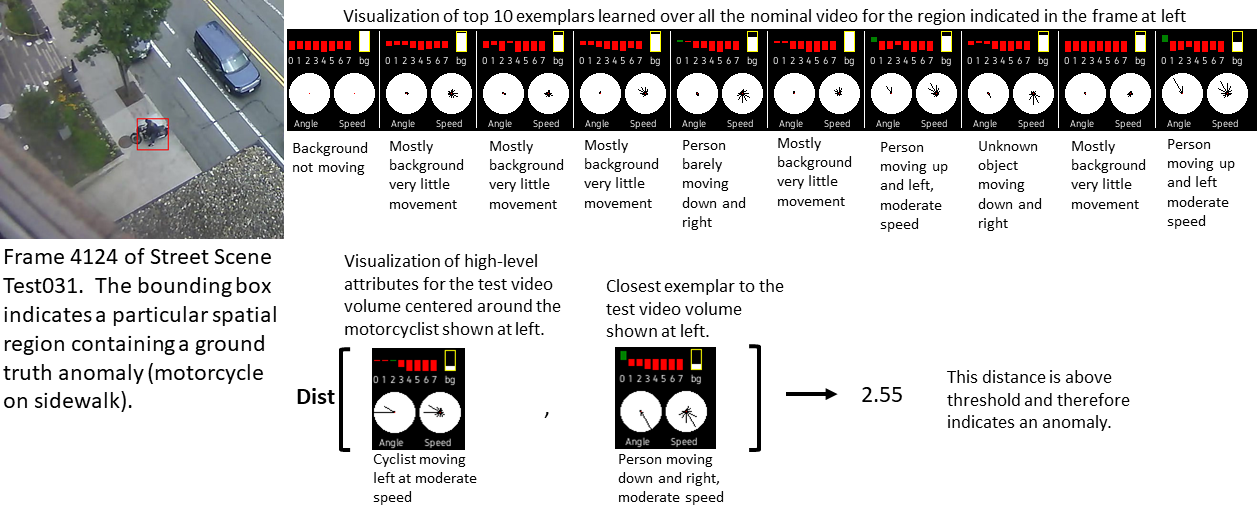}
    \caption{Visualization of the learned exemplars for a region of Street Scene and visualization of a test video volume explaining why it was detected as an anomaly.}
    \label{fig:visualization_motorcycle}
\end{figure*}

Figure \ref{fig:visualization_motorcycle} shows an example from a region on the sidewalk.  The exemplars for this region show either background/unknown objects with very little movement or people/unknown objects moving either up and left or down and right.  The visualization of the high-level attribues estimated for a video volume centered on a person riding a motorcycle onto the sidewalk is shown at the bottom, left of the figure.  It shows that the video volume was estimated to contain a cyclist moving left at high speed.  Although this is not a cyclist, it is a reasonable classification for a motorcyclist.  The nearest exemplar is a person walking down and right at a fast speed.  The distance between the test video volume and the nearest exemplar is large (2.55) and indicates an anomaly.

\begin{figure*}[tbh]
    \centering
    \includegraphics[width=0.8\linewidth]{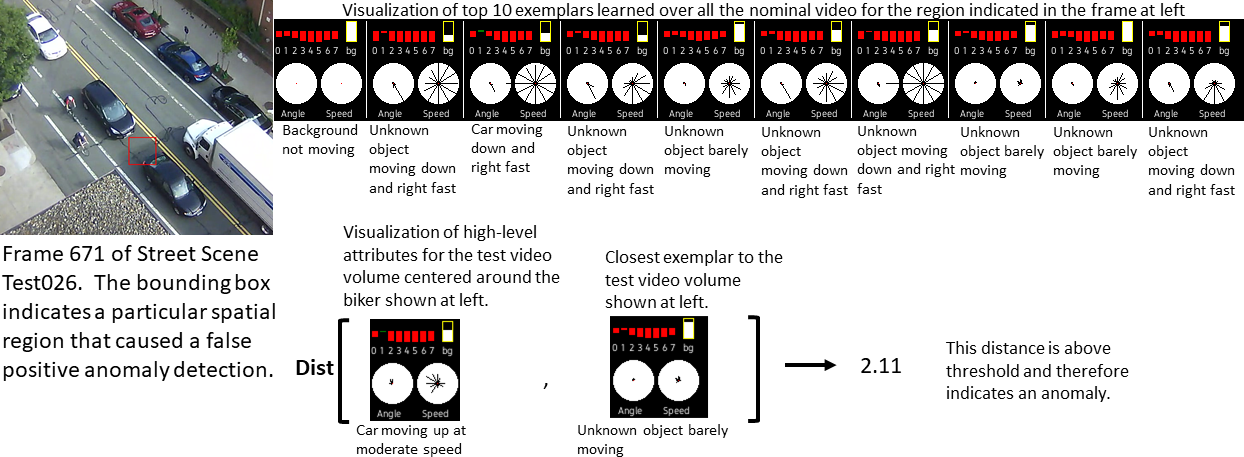}
    \caption{Visualization of the learned exemplars for a region of Street Scene and visualization of a test video volume explaining why it was falsely detected as an anomaly.}
    \label{fig:visualization_falsepos}
\end{figure*}

Figure \ref{fig:visualization_falsepos} shows an example of a false positive anomaly detection in Street Scene.  For the region on the street shown in the frame at the left of the figure, the visualized exemplars show mainly non-moving background/unknown objects or cars/unknown objects moving down and right at various speeds.  The test video volume centered at the frame and region shown at the left of the figure contains the back of a car that is coming to a stop as it moves down and right.  The visualization of this video volume shows that it is estimated by our appearance and motion networks to be a car moving at moderate speed up and left.  This is the opposite direction to how the car is actually travelling and opposite to how cars normally travel in this spatial region.  The angle network has made a mistake in this case.  Thus, the closest exemplar is an unknown object (whose highest likelihood is the car class) barely moving.  Because of the wrongly estimate direction of motion, the anomaly score is high, and an anomaly is falsely indicated.

\begin{figure*}[tbh]
    \centering
    \includegraphics[width=0.8\linewidth]{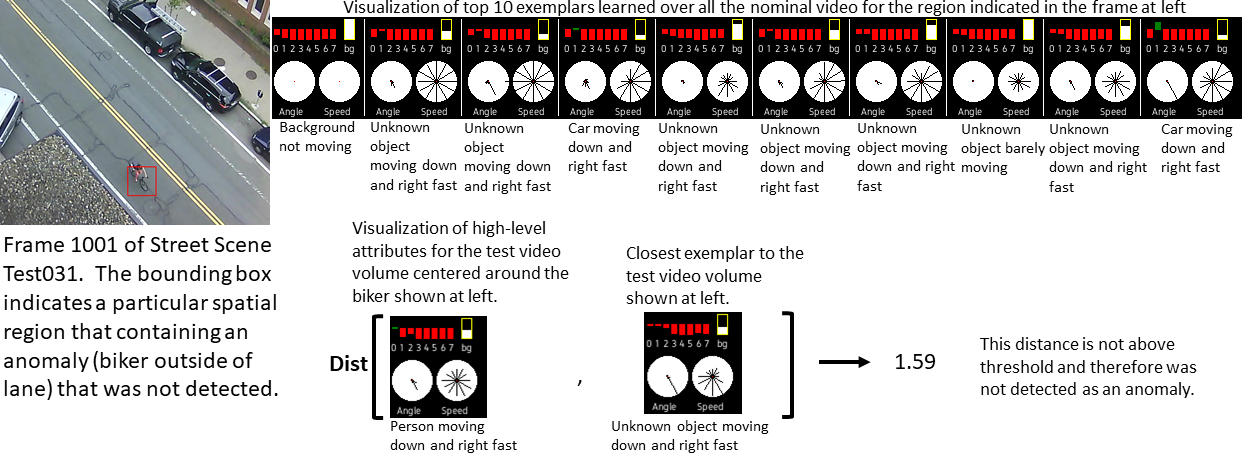}
    \caption{Visualization of the learned exemplars for a region of Street Scene and visualization of a test video volume explaining why it was not detected as an anomaly.}
    \label{fig:visualization_falseneg}
\end{figure*}

As a final example, in Figure \ref{fig:visualization_falseneg} we show a missed anomaly detection on Street Scene.  The region we focus on is on the street and the particular video volume is centered on a cyclist who is outside of the bike lane.  As expected, the exemplars learned for this spatial region show either background/unknown objects with very little movement or cars/unknown objects moving down and right at fast speeds.  The visualization of the video volume containing the anomalous cyclist shows that it was estimated to contain a person moving down and right at a fast speed.  The closest exemplar is an unknown object (although with relatively high likelihoods for person and car) traveling down and to the right at a fast speed.  Because the motion angle and speed match fairly closely and the appearance feature vector is similar, the resulting distance (1.59) is not high enough to indicate an anomaly.  This is mainly a failure of the appearance model to correctly classify the cyclist.

The visualizations of correct anomaly detections as well as false positives and missed detections illustrate how the high-level attributes estimated for each video volume lead to human-understandable explanations of the decisions our system makes.  Analyzing the errors also shows that despite state-of-the-art accuracy on Street Scene, CUHK Avenue and ShanghaiTech datasets, the appearance and motion deep networks are far from perfect and improvements to these networks will directly translate to higher accuracy for video anomaly localization.

\subsection{Limitations}

One general limitation of our approach is that it relies on the appearance and motion networks that estimate high-level features from a video volume.  If these networks are wrong, our method may make a mistaken determination of anomalous/normal, depending on how wrong the networks are.  In general, the more accurate the appearance and motion networks are, the more accurate our anomaly detection method will be.

Our current system has difficulty with a few classes of anomalies in the datasets we have tested on.  On Street Scene, we tend to fail to detect anomalies consisting of cyclists or cars that are slightly outside of their proper lanes.  This could be improved with a finer grid of spatial regions, but at the cost of a higher computational cost.  We also tend to miss very small anomalies in Street Scene (mainly small dogs being walked on the sidewalk).

On the Ped1 and Ped2 datasets, our method has difficulty with skateboarders, especially ones that are traveling about the same speed as pedestrians.  There are often only very subtle motion differences between skateboarders and pedestrians in Ped1 and Ped2 since the skateboard itself is usually barely visible.